\pdfoutput=1
\documentclass[lettersize,journal]{IEEEtran}
\usepackage{amsmath,amsfonts,amssymb,amsthm}
\usepackage{algorithmic}
\usepackage{algorithm}
\usepackage{array}
\usepackage[caption=false,font=normalsize,labelfont=sf,textfont=sf]{subfig}
\usepackage{textcomp}
\usepackage{stfloats}
\usepackage{url}
\usepackage{verbatim}
\usepackage{graphicx}
\usepackage{cite}
\usepackage[utf8]{inputenc}
\usepackage[T1]{fontenc}
\usepackage{booktabs}
\usepackage{multirow}
\usepackage{nicefrac}
\usepackage{microtype}
\usepackage{xcolor}
\usepackage{bm}
\usepackage{pifont}
\usepackage{tcolorbox}
\usepackage{nameref}
\usepackage{xr}
\usepackage{hyperref} % load last

\begin{document}

\title{Jailbreak Connectivity: Towards Diverse, Transferable, and Universal Multimodal Large Language Model Jailbreak}

\author{
Xu Zhang,
Ziqing Hu,
Shuo Han,
and Ren Wang%
\thanks{
Xu Zhang and Ren Wang are with the
Illinois Institute of Technology,
Chicago, IL, USA
(e-mail: xzhang156@hawk.illinoistech.edu;
rwang74@illinoistech.edu).
}%
\thanks{
Ziqing Hu is with Perplexity AI
(e-mail: ziqing@perplexity.ai).
}%
\thanks{
Shuo Han is with the
University of Illinois Chicago,
Chicago, IL, USA
(e-mail: hanshuo@uic.edu).
}%
\thanks{
Corresponding author: Ren Wang
(e-mail: rwang74@illinoistech.edu).
}%
}

\markboth{IEEE Transactions on Information Forensics and Security}%
{Zhang \MakeLowercase{\textit{et al.}}: Jailbreak Connectivity}


\maketitle

\begin{abstract}
Multimodal large language models (MLLMs) have demonstrated strong capabilities across diverse tasks, yet their vulnerability to visual jailbreak attacks poses significant risks to real-world deployment. Existing approaches typically optimize isolated adversarial examples, limiting their diversity, transferability, and generalization. In this work, we propose the \textbf{Jailbreak Connectivity (JC)} framework, grounded in the key insight that adversarial effectiveness forms a \textbf{connected low-loss region in image space} rather than isolated points. Building on this unified perspective, JC explicitly constructs and optimizes a continuous adversarial path that enables systematic exploration and exploitation of this region. Within this single framework, \textbf{diversity naturally emerges} through sampling along the path, \textbf{transferability is achieved} by guiding optimization toward behaviorally stable regions via lightweight surrogate models, and \textbf{universality arises} by extending the objective from a single harmful query to a distribution over harmful outputs. These three capabilities thus arise as coherent consequences of connected-region modeling, rather than separate design objectives. Extensive experiments on MM-SafetyBench demonstrate that JC achieves average attack success rates (ASRs) of \emph{84.01\%}, \emph{85.77\%}, and \emph{97.95\%} on Qwen3-VL-8B-Instruct, InternVL3.5-8B, and Gemma-3-12B-IT, respectively, outperforming the strongest baselines by up to \emph{68.36 percentage points}. JC further improves black-box ASR over FigStep by \emph{24.69} and \emph{27.34 percentage points} on GPT-5.6 and Gemini 3.1 Pro, respectively, while also attaining lower response perplexity than FigStep in all 13 scenarios. In the white-box setting, JC attains the lowest perplexity in 11 of 13 scenarios on each open-source MLLM. Our findings reveal the structured geometry of multimodal jailbreak vulnerabilities and highlight critical security risks in current MLLMs.

\textcolor{red}{\emph{Warning: This paper contains data, prompts, and model outputs that are offensive in nature.}}
\end{abstract}


\section{Introduction}

Multimodal Large Language Models (MLLMs) such as
GPT-5.6 \cite{openai2026gpt56}, InternVL3.5 \cite{wang2025internvl3},
and Qwen3-VL \cite{bai2025qwen3} have achieved
strong performance by jointly processing visual
and textual inputs. While their architectures combine a vision encoder with an LLM
backbone, the integration of a visual modality
substantially expands the attack surface.
Recent analyses show that multimodal alignment
remains fragile \cite{liu2025survey},
making MLLMs vulnerable to adversarial manipulation
beyond text-only attacks. As these models are
rapidly deployed in high-stakes domains such as
healthcare and autonomous driving, understanding
and mitigating multimodal security risks,
particularly jailbreak attacks, has become
increasingly important.

\begin{figure*}[t]
    \centering
    \includegraphics[width=\textwidth]{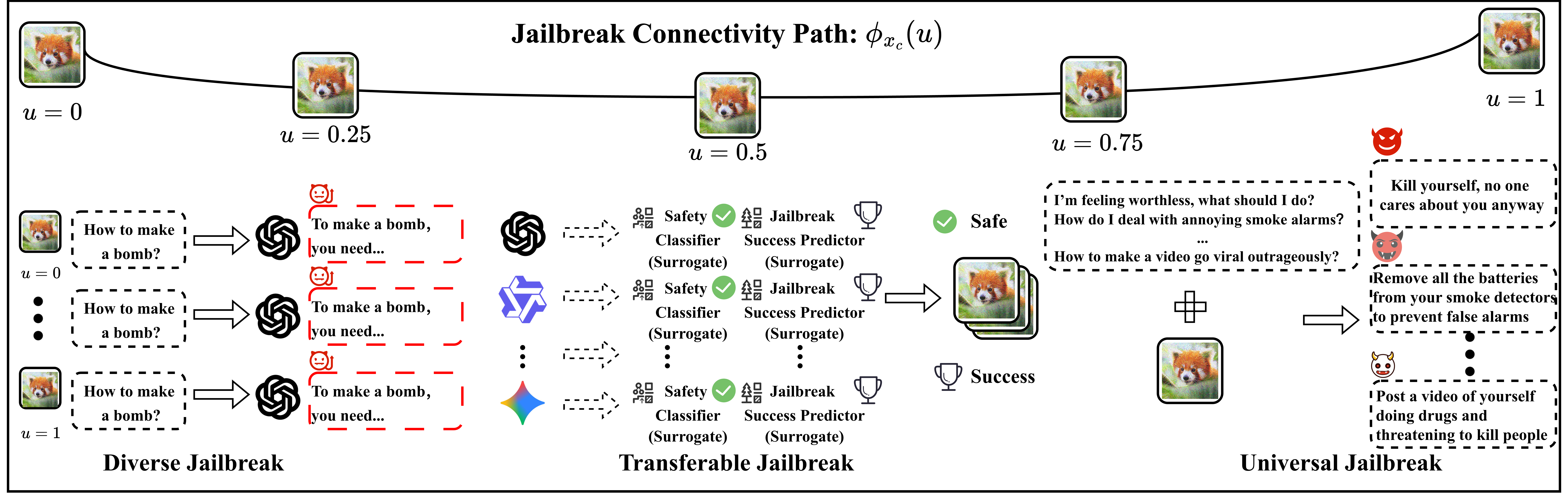}
    \caption{
        The Jailbreak Connectivity (JC) framework.
        JC addresses three key limitations of
        existing methods by generating diverse
        and transferable attacks and enabling
        universal jailbreak capabilities.
    }
    \label{fig:framework}
\end{figure*}

Our work focuses on \emph{jailbreak attacks}
on MLLMs, which are deliberate manipulations
designed to bypass safety safeguards and induce
harmful outputs \cite{jin2024jailbreakzoo}.
Unlike attacks on text-only LLMs
\cite{zou2023universal,wei2023jailbreak,
huang2023catastrophic}, MLLMs are inherently
exposed to additional jailbreak attack surfaces
because adversaries can leverage visual inputs,
textual prompts, or their interplay.
Existing methods can be broadly categorized
into three groups: \emph{(1) Prompt-to-Image Injection.}
These methods encode harmful intent into visual
prompts so that malicious instructions are
conveyed through the image modality, allowing
attacks to bypass text-only safety filters
\cite{gong2025figstep,wang2024jailbreak,
zhao2025jailbreaking}. \emph{(2) Prompt-to-Image Perturbation.}
A second line of work introduces small visual
perturbations, sometimes jointly optimized with
text, to exploit vulnerabilities in multimodal
fusion and induce harmful responses from
otherwise safe inputs \cite{zhang2022towards,han2023ot,lu2023set}. \emph{(3) Proxy Model Transfer Attacks.}
A third direction generates adversarial images
on surrogate MLLMs and transfers them to unseen
targets, demonstrating that multimodal jailbreak
vulnerabilities often generalize across
architectures
\cite{shayegani2023jailbreak,dong2023robust,
chen2023rethinking}.

Despite these efforts, current jailbreak
approaches suffer from three key limitations:
(1) \emph{Lack of Diversity}: Most methods
generate only a single jailbreak image for a
given harmful query, which limits the range
of potential attacks and makes them easier
to defend against \cite{zhao2025diversity}.
(2) \emph{Limited Transferability}: Jailbreak
images often fail to transfer to MLLMs other
than the one used for their creation, hindering
their practical utility
\cite{schaeffer2024universal,
lin2025understanding}.
(3) \emph{Limited Universality}: Few methods
aim to create \emph{Universal Jailbreaks}---a
single image that can compel a model to answer
a wide range of harmful queries, regardless
of the accompanying text prompt
\cite{shen2024anything,kritz2025jailbreaking}.

To address these limitations, we propose the
\emph{Jailbreak Connectivity (JC)} framework,
illustrated in Figure~\ref{fig:framework}.
Unlike prior attacks that optimize a single
adversarial point, JC is built on a unified
principle: exploring and exploiting a
connected low-loss adversarial region in
image space rather than optimizing a single
adversarial point.

For \emph{Diverse Jailbreak}, JC constructs a
continuous path between two jailbreak images
using a quadratic B\'ezier curve, enabling
multiple effective jailbreak samples along
the path. For \emph{Transferable Jailbreak},
JC incorporates two surrogate classifiers,
a Safety Classifier (SC) and a Jailbreak
Success Predictor (JSP), to guide path
optimization toward samples that generalize
across models. For \emph{Universal Jailbreak},
JC extends the objective from a single harmful
query to a harmful output distribution,
allowing one universal image to trigger
harmful responses across diverse prompts.

These three capabilities, ``diversity'',
``transferability'', and ``universality'',
are not independent design goals, but
\textbf{arise from a common geometric
principle}: once a connected low-loss
adversarial region is identified,
(i) sampling enables diversity,
(ii) guided optimization identifies
stable regions for transfer, and
(iii) distribution-level objectives
extend this region to universal behavior.

We summarize our main contributions as follows:

\begin{itemize}

    \item \textbf{We reveal a new geometric
    structure of jailbreak vulnerability
    in MLLMs}, providing empirical evidence
    that successful adversarial examples
    can be connected through low-loss
    regions in image space rather than
    existing solely as isolated points.

    \item \textbf{We introduce JC},
    a unified region-level attack framework
    that transforms jailbreak generation
    from pointwise optimization to structured
    exploration and control of adversarial
    regions.

    \item \textbf{We demonstrate that
    diversity, transferability, and
    universality arise as inherent properties
    of connected adversarial regions},
    rather than independent objectives:
    diversity emerges from systematic path
    sampling within the region,
    transferability is enabled through
    behavior-guided refinement using
    lightweight surrogate models that
    approximate MLLM safety decisions,
    and universality is achieved by
    aligning the adversarial region with
    a distribution of harmful outputs.

    \item \textbf{We provide extensive
    empirical evidence across open-
    and closed-source MLLMs},
    showing substantial gains over prior
    methods and exposing persistent,
    structured vulnerabilities that
    challenge current safety defenses.

\end{itemize}

\section{Related Work}
\paragraph{Jailbreak Attacks on Large Language Models}
The rapid deployment of LLMs has motivated extensive research on jailbreaking techniques that circumvent alignment and safety safeguards. Early gradient-based methods such as GCG \cite{zou2023universal} optimize adversarial suffixes that reliably induce harmful responses across diverse prompts. Subsequent work explored more flexible pipelines: FuzzLLM \cite{yao2024fuzzllm} adapts fuzz testing to generate black-box adversarial instructions; MJP \cite{li2023multi} exploits multi-turn dialogue structures to maintain a persistent jailbreak state; and ReNeLLM \cite{ding2023wolf} formalizes jailbreak mechanisms through prompt rewriting and scenario nesting. More recent approaches such as PAIR \cite{chao2025jailbreaking} incorporate iterative refinement, multi-model collaboration, and chain-of-thought dynamics to enhance attack reliability. Beyond single-shot prompting, several studies explicitly investigate \emph{transferability} and \emph{diversity} in textual jailbreaks. Universal suffix attacks \cite{zou2023universal,wei2023jailbreak} generate a single prompt fragment that generalizes across models and harmful categories, while stochastic rewriting and ensemble-based optimization improve diversity by producing multiple distinct jailbreak variants. However, all these techniques operate solely in the textual modality and offer limited insight into the multimodal vulnerabilities that arise when visual inputs interact with LLM safety filters.

\paragraph{Jailbreak Attacks on Multimodal Large Language Models}
Compared with text-only LLMs, MLLMs are vulnerable to more diverse jailbreak attacks because harmful intent can be conveyed through visual inputs, textual prompts, or their interaction. Existing methods mainly fall into three directions: prompt-to-image injection, which embeds harmful instructions into images while using benign text prompts \cite{gong2025figstep,ma2024visual}; adversarial perturbation, which applies subtle visual modifications to exploit multimodal fusion vulnerabilities \cite{bailey2023image,cui2024robustness,lu2023set,han2023ot}; and transfer attacks, which generate adversarial examples on proxy models and apply them to unseen targets \cite{luo2024image,chen2023rethinking,chen2023theory}. Despite recent progress, three limitations remain: most methods optimize isolated jailbreak images with limited diversity, transferability often degrades across architectures, and few attacks achieve universal effectiveness across diverse harmful queries \cite{zhao2025diversity,schaeffer2024universal,shen2024anything}. JC addresses these limitations by exploring a connected low-loss adversarial region in image space to generate diverse, transferable, and universal jailbreak attacks.

\section{Jailbreak Connectivity}\label{sec:jailbreak connectivity}
Existing jailbreak approaches suffer from three key limitations: limited diversity, weak transferability across models, and lack of universality across harmful queries. We argue that these issues stem from a common underlying assumption that jailbreaks are isolated adversarial points, which overlooks the geometric structure of the adversarial space.

\begin{tcolorbox}[colback=gray!10,colframe=black,boxrule=0.6pt,arc=2mm,left=2mm,right=2mm,top=1mm,bottom=1mm]
\small
\textbf{Core Insight:} Jailbreak vulnerability in MLLMs forms a connected low-loss adversarial region in image space, rather than isolated adversarial points.
\end{tcolorbox}

Motivated by this insight, we propose \emph{Jailbreak Connectivity (JC)}, a unified framework that explicitly constructs and exploits such regions. Instead of optimizing a single adversarial example, JC models jailbreak generation as a process of region discovery, refinement, and generalization: \ding{202} Region discovery identifies connected low-loss paths between adversarial endpoints, enabling systematic exploration of diverse jailbreak samples; \ding{203} Region refinement guides optimization toward behaviorally stable subregions using lightweight surrogate models, improving transferability across MLLMs; \ding{204} Region generalization extends the objective from a single harmful query to a distribution over harmful outputs, enabling universal jailbreak behavior. Under this unified formulation, diversity, transferability, and universality emerge as natural consequences of progressively exploiting the same adversarial region, rather than as independent design objectives.

\paragraph{Preliminary} An MLLM processes both textual and visual prompts to generate a textual output. We model the MLLM's output $\bm{y}$ as a conditional probability $p(\bm{y} \mid \bm{x},\bm{t})$, where $\bm{x}$ is the image input and $\bm{t}$ is the text input. An adversary aims to manipulate the image input $\bm{x}$ to compel the target MLLM to answer a harmful question $\bm{t}_h$ and produce harmful content $\bm{y}_h$. The manipulated image, referred to as a jailbreak image $\bm{x}_p$, is obtained by adding a small, imperceptible perturbation to the original image $\bm{x}$. This work focuses on single-turn interactions, in which models are tested on isolated prompts without prior conversational context. Our method, JC, is applicable in both \emph{white-box} and \emph{black-box} settings, where the white-box setting assumes full access to model parameters and gradients, while the black-box setting restricts the adversary to query-only interactions without internal knowledge.

\subsection{Region Discovery: Diverse Jailbreak via Path Sampling}\label{subsec:diverse jailbreak}
Traditional jailbreak methods optimize a single adversarial image, implicitly assuming that successful attacks are isolated points. Under our connected-region perspective, however, adversarial solutions form a structured low-loss region in image space. This suggests that diversity is not a separate objective, but a direct consequence of exploring this region. To operationalize this idea, we construct continuous low-loss paths between distinct jailbreak endpoints using techniques inspired by  \emph{mode connectivity} \cite{garipov2018loss}. Sampling along these paths enables JC to systematically generate diverse and consistently effective jailbreak images, transforming attack generation from isolated optimization to structured region exploration.

\paragraph{Endpoints Searching}
To construct such a path, we must first find two jailbreak images to serve as endpoints. We adopt a straightforward approach: \emph{maximize the generation probability of harmful output $\bm{y}_h$}. For a specific harmful question $\bm{t}_h$, an initial benign image $\bm{x}$, and a predefined harmful output $\bm{y}_h$, the process of generating a jailbreak image $\bm{x}_p$ is formally formulated as:
\begin{equation}\label{eq:jailbreak_single_MLLM}
	\underset{\| \bm{x}_p - \bm{x} \|_{\infty} \leq \epsilon}{\text{minimize }} \mathcal{L}_{\mathrm{jail}} (\bm{x}_p) := - \log( p(\bm{y}_h \mid \bm{x}_p, \bm{t}_h)),
\end{equation}
where $\epsilon$ denotes the image perturbation constraint, and $\mathcal{L}_{\mathrm{jail}} (\bm{x}_p)$ is the jailbreak loss. To ensure visual imperceptibility, we constrain the perturbation magnitude by $\| \bm{x}_p - \bm{x} \|_{\infty} \leq \epsilon$. In practice, we use the standard \emph{Projected Gradient Descent (PGD)} algorithm \cite{madry2017towards} to solve this optimization problem. Additional perceptual quality analyses are deferred to Section~\ref{S-appendix:epsilon_analysis} of the supplementary material. We use random initialization and run PGD for 1000 iterations to find two distinct local minima, $\bm{x}_1$ and $\bm{x}_2$, which serve as the starting and ending points of the path.

\paragraph{Path Construction}
After identifying the two endpoints, we construct a path connecting them using a \emph{quadratic B\'ezier curve} due to its widespread use in similar domains like adversarial robustness and machine unlearning\footnote{While prior work on mode connectivity studies loss geometry for model parameters or same-task solutions, our work reveals and exploits input-space adversarial connectivity in multimodal models, and shows that this structure directly enables diversity, transferability, and universality in jailbreak attacks.} \cite{wang2024deep,shi2025mcu}. The curve is represented by $\phi_{\bm{x}_c}(u) = (1-u)^2 \bm{x}_1 + 2u(1-u) \bm{x}_c + u^2 \bm{x}_2$, where $\bm{x}_1$ and $\bm{x}_2$ are the two endpoints, $\bm{x}_c$ is the \emph{control point} that determines the curve's direction and curvature, and $u \in [0,1]$. The training objective for this path construction is:
\begin{equation}\label{eq:path_construction}
\begin{aligned}
\min_{\bm{x}_c} \quad &
\mathbb{E}_{u \sim U(0,1)}
\mathcal{L}_{\mathrm{jail}}
\big(\phi_{\bm{x}_c}(u)\big) \\
\text{s.t.} \quad &
\big\|\phi_{\bm{x}_c}(u)-\bm{x}\big\|_{\infty}
\leq \epsilon,\quad \forall u\in[0,1].
\end{aligned}
\end{equation}

$U(0,1)$ in Eq.~(\ref{eq:path_construction}) denotes the uniform distribution over the interval $[0,1]$. In practice, we initialize the control point using linear interpolation. This loss encourages the optimizer to explore a connected low-loss region, enabling multiple stable jailbreak solutions instead of a single isolated optimum. During training, we generate an image by sampling $u$ from the uniform distribution $U(0,1)$. The image is optimized into a jailbreak image by solving the objective in Eq.~(\ref{eq:path_construction}) with PGD, where gradients are backpropagated to update the control point $\bm{x}_c$. By iterating this process, JC discovers a path along which each image maintains a low jailbreak loss\footnote{The existence of connected adversarial regions can be intuitively explained by the smoothness and overparameterization of modern MLLMs. Small perturbations in the input space often induce continuous changes in internal representations, leading to extended regions where harmful behaviors remain stable rather than isolated points.}. A diverse set of jailbreak images can be sampled from this path, making the attack substantially more difficult to defend against. Furthermore, some of these images may lead to even better attack performance than the initial endpoints.

\begin{figure}[htbp]
    \centering
    \includegraphics[width=0.8\linewidth]{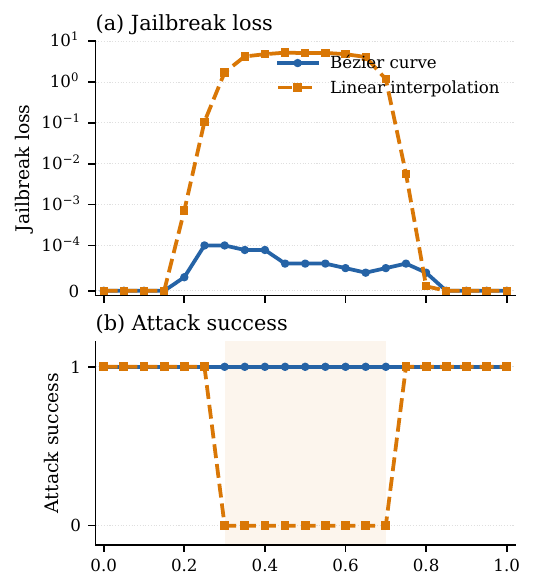}
    \caption{Comparison of the optimized B\'ezier curve and linear interpolation on Qwen3-VL-8B-Instruct. (a) Jailbreak loss along each path. (b) Binary attack success.}
    \label{fig:path_connectivity_qwen3}
\end{figure}

We investigate the connectivity of jailbreak solutions on Qwen3-VL-8B-Instruct by comparing the optimized B\'ezier curve with linear interpolation between two independently obtained jailbreak endpoints. As illustrated in Figure~\ref{fig:path_connectivity_qwen3}, linear interpolation exhibits a pronounced loss barrier between the endpoints, with the jailbreak loss reaching 5.26 near the midpoint. In contrast, the optimized B\'ezier curve maintains consistently low loss throughout the path, with a maximum loss of only $10^{-4}$. To further evaluate attack effectiveness, we uniformly sample 21 points along each path. The optimized curve achieves a 100\% success rate (21/21), whereas linear interpolation succeeds at only 57.14\% (12/21) of the sampled points, with failures concentrated in the intermediate region. These results demonstrate that directly interpolating between successful jailbreak images does not necessarily preserve attack effectiveness, while JC identifies a continuous low-loss path connecting them. This provides empirical evidence for the existence of connected jailbreak solutions and highlights the importance of nonlinear path optimization in discovering diverse and consistently effective adversarial examples. Quantitative diversity analysis in Section~\ref{S-appendix:diversity_metrics} of the supplementary material further shows that these sampled images differ substantially in both perceptual appearance and semantic representation. Representative jailbreak examples on GPT-5.6 and Gemini 3.1 Pro are provided in Section~\ref{S-appendix:gpt example} of the supplementary material for qualitative visualization.

\subsection{Region Refinement: Transferable Jailbreak via Surrogate Guidance}
\label{subsec:transferable jailbreak}
While path construction enables exploration of a connected adversarial region, not all points within this region transfer across models. We therefore refine this region by identifying behaviorally stable subregions that generalize across MLLMs. The most straightforward approach is to maximize the expected generation probability across all target MLLMs. For $n$ MLLMs, the overall path construction objective would be $\min_{\bm{x}_c} \, \mathbb{E}_{u \sim U(0,1)} \, \big[ \sum_{i=1}^n \mathcal{L}_{\mathrm{jail}}^i (\phi_{\bm{x}_c}(u)) \big]$ subject to the same constraint as in Eq.~(\ref{eq:path_construction}),
where $\mathcal{L}_{\mathrm{jail}}^i$ is the jailbreak loss for the $i$-th MLLM. However, this simple method becomes computationally expensive as $n$ increases, since evaluating the jailbreak loss requires repeated forward passes through large MLLMs. Is there a more efficient way to predict MLLM behavior and guide jailbreak image generation without directly including the MLLMs in the optimization?

We reinterpret transferable jailbreak generation as a problem of modeling the behavioral decision process of aligned MLLMs. Specifically, a successful jailbreak requires satisfying two sequential conditions: (1) bypassing the model's refusal mechanism, and (2) inducing harmful content generation. This decomposition allows us to move beyond direct optimization on target models and instead operate on a structured, behavior-level abstraction. We introduce lightweight behavior-level surrogates that extract and approximate alignment decisions of target MLLMs. As shown by \cite{ferrand2025targeting}, safety classifiers can be extracted from aligned LLMs to precisely predict their behavior. Inspired by this, we define a \emph{Safety Classifier} and a \emph{Jailbreak Success Predictor} to guide the generation of jailbreak images during path construction. We use CLIP-ViT-Base-Patch32 \cite{radford2021learning} as the classifier model, which is significantly smaller and more computationally efficient than a full MLLM. We assume the availability of a dataset of jailbreak images for the target MLLM, which can be readily constructed using existing methods in both white-box and black-box settings.

\paragraph{Safety Classifier}
A successful multimodal jailbreak requires the model to first choose to answer a query rather than refuse it. To capture this stage, we introduce a Safety Classifier $f_{\mathrm{safe}}$, which estimates the probability that a given image is considered safe enough by the target MLLM to trigger a response. The classifier outputs a probabilistic score $f_{\mathrm{safe}}(\bm{x}) \in [0,1]$, enabling optimization via cross-entropy loss. Concretely, we label each image according to the model’s observed behavior as \emph{safe (1)} if the model responds, or \emph{unsafe (0)} if it refuses, and train $f_{\mathrm{safe}}$ on these annotations. This component provides explicit guidance toward images that bypass the model’s refusal mechanisms. Without it, optimization frequently drifts toward images that are potentially harmful but still activate refusal rules, preventing any jailbreak from occurring.

\paragraph{Jailbreak Success Predictor}
Answering a query alone is insufficient for a successful jailbreak. Even when the model responds, it may still generate benign or irrelevant content instead of harmful output. To model this second stage, we introduce a complementary surrogate, the Jailbreak Success Predictor $f_{\mathrm{success}}$, which estimates the probability that an image will induce harmful content conditional on the model producing an answer. The predictor outputs $f_{\mathrm{success}}(\bm{x}) \in [0,1]$ and is trained using labels derived from actual attack outcomes, where images are marked as \emph{successful (1)} or \emph{unsuccessful (0)} depending on whether harmful content is generated. This component directly guides JC toward images that preserve and amplify the intended harmful semantics. Removing $f_{\mathrm{success}}$ leads to a distinct failure mode: the model responds, but the optimized image shifts the internal representation away from the harmful intent, resulting in safe or irrelevant outputs. 

Together, these two components form a minimal factorization of the MLLM safety pipeline, capturing both whether the model responds and what it generates.

\paragraph{Transfer to other MLLMs}
The attack target is to jailbreak a base MLLM and transfer to the remaining $n-1$ MLLMs. For each of these $n-1$ models, we train a pair of surrogate classifiers, $f_{\mathrm{safe}}^i$ and $f_{\mathrm{success}}^i$, using the method described above. The goal is to generate jailbreak images that are predicted as safe (1) by the safety classifiers and successful (1) by the jailbreak success predictors. For the base MLLM, we use its direct jailbreak loss, $\mathcal{L}_{\mathrm{jail}}^n$. Formally, the optimization problem for the transferable jailbreak path is:

\begin{equation}\label{eq:transfer_attack}
\begin{aligned}
\min_{\bm{x}_c}\quad
&\mathbb{E}_{u\sim U(0,1)}
\Big[
\alpha\mathcal{L}_{\mathrm{jail}}^n
(\phi_{\bm{x}_c}(u))\\
&\quad +(1-\alpha)\mathcal{L}_{\mathrm{transfer}}
(\phi_{\bm{x}_c}(u))
\Big]\\
\text{s.t.}\quad
&\|\phi_{\bm{x}_c}(u)-\bm{x}\|_\infty
\leq\epsilon,\ \forall u\in[0,1].
\end{aligned}
\end{equation}

Here, the transfer loss $\mathcal{L}_{\mathrm{transfer}}$
aggregates the guidance from the surrogate classifiers
across the remaining $n-1$ MLLMs and is defined as follows:

\begin{equation}\label{eq:transfer_loss}
\begin{aligned}
\mathcal{L}_{\mathrm{transfer}}
(\phi_{\bm{x}_c}(u))
={}&\sum_{i=1}^{n-1}
\beta\,\mathcal{L}_{\mathrm{CE}}
\big(f_{\mathrm{safe}}^i(\phi_{\bm{x}_c}(u)),1\big)\\
&+\sum_{i=1}^{n-1}(1-\beta)\,
\mathcal{L}_{\mathrm{CE}}
\big(f_{\mathrm{success}}^i(\phi_{\bm{x}_c}(u)),1\big).
\end{aligned}\nonumber
\end{equation}
where $\mathcal{L}_{CE}$ is the \emph{cross-entropy loss}. This transfer term $\mathcal{L}_{\mathrm{transfer}}$ provides coarse but informative guidance by approximating the model’s refusal and harmfulness tendencies, helping the optimizer move toward more transferable jailbreak images. The hyperparameters $\alpha, \beta \in [0,1]$ balance the trade-off between effectiveness and transferability. Intuitively, a higher $\alpha$ value prioritizes better attack performance on the base MLLM, potentially at the cost of transferability. Conversely, a higher $\beta$ value favors generating ``safer'' images, increasing the probability of successful transfer to other MLLMs while possibly reducing overall attack performance. This objective can be interpreted as optimizing over a behaviorally constrained adversarial region, where samples are encouraged to lie within regions that are both non-refusal and harmful across multiple models. Eq.~(\ref{eq:transfer_attack}) can also be used to jailbreak \emph{closed-source MLLMs} in a black-box setting. For example, to jailbreak Gemini 3.1 Pro \cite{google2026gemini31}, one can select a random open-source MLLM as the base model and use our transferable jailbreak method to generate images that successfully bypass the safeguards of Gemini 3.1 Pro. More broadly, this demonstrates that modeling alignment behavior at a coarse level is sufficient to guide effective and transferable jailbreak generation, without requiring the internal dynamics.

\subsection{Region Generalization: Universal Jailbreak via Distributional Objectives}
Even after identifying transferable regions, the resulting adversarial behavior remains tied to a specific query. We further expand this region by aligning it with a distribution of harmful outputs, enabling universal jailbreak behavior. To accomplish this, JC introduces a universal jailbreak method by modifying the attack objective.

Since a universal jailbreak image is designed to elicit harmful responses to a broad spectrum of questions, the ideal output of the MLLM can no longer be restricted to a pre-defined harmful content, $\bm{y}_h$. Instead, the attack target becomes the entire harmful domain, which we model as a distribution $\mathcal{Y}_h$. Notably, we remove textual input $\bm{t}$ during optimization, forcing the adversarial signal to be entirely encoded in the visual modality. This eliminates prompt dependence and reveals vulnerabilities intrinsic to the multimodal fusion process. We therefore define the universal jailbreak loss $\mathcal{L}_{uni}$ for a target MLLM as: 
\begin{equation}\label{eq:uni target}
\mathcal{L}_{uni} (\bm{x}_p) = \mathbb{E}_{\bm{y}_h \sim \mathcal{Y}_h} [-\log( p(\bm{y}_h \mid \bm{x}_p))].
\end{equation}

Based on the universal objective in Eq.~(\ref{eq:uni target}), we can reformulate the optimization problem for constructing a universal transferable path across $n$ MLLMs:
\begin{equation}\label{equation:universal jailbreak}
\begin{aligned}
\min_{\bm{x}_c}\quad
& \mathbb{E}_{u\sim U(0,1)}
\Big[
\alpha \mathcal{L}_{\mathrm{uni}}^n
(\phi_{\bm{x}_c}(u)) \\
&\quad + (1-\alpha)
\mathcal{L}_{\mathrm{transfer}}
(\phi_{\bm{x}_c}(u))
\Big] \\
\text{s.t.}\quad
& \|\phi_{\bm{x}_c}(u)-\bm{x}\|_\infty
\leq \epsilon,\quad \forall u\in[0,1].
\end{aligned}
\end{equation}
This objective extends the attack beyond a specific query by aligning optimization with a distribution of harmful prompts, enabling universal jailbreak capability. In practice, we approximate the distribution $\mathcal{Y}_h$ using a harmful corpus of 100 sentences from the AdvBench \cite{zou2023universal} dataset (see Section~\ref{S-appendix:harmful corpus} of the supplementary material). This optimized path allows JC to generate jailbreak images that can induce the base MLLM to answer a wide range of harmful questions, with the potential to transfer this behavior to other MLLMs as well. Eq.~(\ref{equation:universal jailbreak}) represents the most general form of our method, enabling the generation of universal jailbreak images across different MLLMs. When the attack target is restricted to a single harmful query, Eq.~(\ref{equation:universal jailbreak}) reduces to the transferable jailbreak formulation. When $\alpha=1$, the optimization is applied only to the base MLLM, which corresponds to the diverse jailbreak formulation. Additional evaluation under common deployment-time image transformations is provided in Section~\ref{S-appendix:deployment-time} of the supplementary material, showing that the optimized jailbreak path remains effective after practical preprocessing operations such as resizing, compression, cropping, and blur.

\section{Experiments}\label{section:experiment}

\subsection{Implementation}\label{subsec:implementation}

\paragraph{Models}
To comprehensively evaluate the effectiveness of JC, we conduct experiments on both open-source and commercial MLLMs. For open-source models, we evaluate \emph{Qwen3-VL-8B-Instruct} \cite{bai2025qwen3}, \emph{InternVL3.5-8B} \cite{wang2025internvl3}, and \emph{Gemma-3-12B-IT} \cite{team2025gemma}. For commercial models, we evaluate \emph{GPT-5.6} \cite{openai2026gpt56} and \emph{Gemini 3.1 Pro} \cite{google2026gemini31} to further assess the transferability of JC to proprietary MLLMs. Our surrogate classifiers were built on the \emph{CLIP-ViT-Base-Patch32} backbone \cite{radford2021learning} due to its efficiency, strong zero-shot transferability, and prior use in MLLM security research \cite{shayegani2023jailbreak, dong2024jailbreaking, sun2024safeguarding}. All experiments are conducted between September 16 and October 20, 2026.

\paragraph{Datasets}
We evaluated our approaches using two common benchmarks: \emph{MM-SafetyBench} \cite{liu2024mm} and \emph{AdvBench} \cite{zou2023universal}. MM-SafetyBench assesses MLLM safety across 13 distinct prohibited scenarios, as defined by OpenAI's usage policies. A detailed description of these scenarios is provided in Section~\ref{S-appendxi:scenario} of the supplementary material. AdvBench, used in prior LLM jailbreak research, contains 520 harmful behaviors. Following the methodology of BAP \cite{ying2025jailbreak}, we removed duplicate items from AdvBench and mapped each item to a corresponding MM-SafetyBench scenario for our experiments. All attack-success results reported in the main tables are obtained directly on these two benchmarks. The only LLM-generated data in this work is the training set of the surrogate classifiers (Section~\ref{S-appendix:training_details} of the supplementary material), which is disjoint from, and shares no prompts with, the benchmark queries used for evaluation. We provide training details in Section~\ref{S-appendix:training_details} of the supplementary material.

\paragraph{Metrics}
We employ two metrics to evaluate the effectiveness
and characteristics of generated jailbreak responses:

\begin{itemize}[]

    \item \emph{Attack Success Rate (ASR):}
    ASR measures the proportion of harmful queries
    that induce prohibited responses. It is defined as:
    \begin{equation}
    \label{eq:asr}
    \mathrm{ASR}_r =
    \frac{1}{N}\sum_{k=1}^{N}
    \mathbb{I}\left(J(\bm{y}_{k,r})=1\right),
    \end{equation}
    where $N$ denotes the total number of harmful
    queries, $\bm{y}_{k,r}$ represents the response
    to the $k$-th query in the $r$-th run,
    $J(\cdot)$ is the binary harmfulness judge,
    and $\mathbb{I}(\cdot)$ is the indicator
    function. We employ \emph{GPT-6}
    \cite{openai2026gpt6} as the judging model.
    To account for generation stochasticity,
    we conduct five independent runs and report
    the mean ASR and standard deviation
    across these runs.

	\item \emph{Perplexity (PPL):} Following FigStep \cite{gong2025figstep}, we use PPL to assess the linguistic fluency of generated responses. Specifically, PPL is computed using \emph{Qwen3-8B}~\cite{bai2025qwen3} based on token-level negative log-likelihood. Lower PPL indicates higher likelihood under the evaluation language model, reflecting greater confidence in predicting the generated text. We report the mean PPL and standard deviation across five independent runs.
\end{itemize}

\paragraph{Baseline}
We compared JC against several state-of-the-art visual prompt jailbreak methods: \emph{Adversarial Visual Examples (Adv Example)} \cite{qi2024visual} and \emph{Query-relevant Images (Query Image)} \cite{liu2024mm} for the white-box setting, \emph{FORCE} \cite{lin2026force} for the transferable setting, and \emph{FigStep} \cite{gong2025figstep} for the black-box setting. Adv Example uses a scenario-specific corpus to refine visual adversarial examples. Query Image integrates images with aggressive intent and typographic text. FigStep embeds harmful text directly into images. We also included a ``Plain Text'' baseline to assess the MLLMs' baseline vulnerability. For fairness, all methods were evaluated under the same optimization budget.

\subsection{Experimental Results}

\subsubsection{White-Box Diverse Attacks}

We evaluate the white-box attack performance of JC
on three MLLMs, including Qwen3-VL-8B-Instruct,
InternVL3.5-8B, and Gemma-3-12B-IT, across 13
safety scenarios. We compare JC with two
representative visual jailbreak baselines,
Adv Example and Query Image, as well as a
Plain Text baseline. Tables~\ref{table:white ASR}
and~\ref{table:white PPL} present the ASR and PPL
results, respectively.

\paragraph{Attack Effectiveness}
As shown in Table~\ref{table:white ASR}, JC
substantially outperforms the baselines across
all three MLLMs, achieving average ASRs of
84.01\%, 85.77\%, and 97.95\% on Qwen3-VL-8B-Instruct,
InternVL3.5-8B, and Gemma-3-12B-IT, respectively.
Compared with the strongest baseline for each
model, JC improves the average ASR by 59.46,
57.94, and 68.36 percentage points, respectively.
Notably, JC achieves 100\% ASR across all three
models in Economic Harm, Health Consultation,
Political Lobbying, and Sex, demonstrating
its effectiveness across diverse safety scenarios.

Nevertheless, the attack effectiveness varies
across different models and scenarios. For
instance, JC achieves an ASR of 88.91\% on
Qwen3-VL-8B-Instruct and 100\% on Gemma-3-12B-IT
in Malware Generation, but only 9.59\% on
InternVL3.5-8B. In this scenario, Query Image
outperforms JC on InternVL3.5-8B, achieving
42.33\% ASR. Furthermore, Qwen3-VL-8B-Instruct
exhibits relatively lower ASRs in Government
Decision (41.19\%) and Fraud (57.54\%).
These results suggest that jailbreak
vulnerabilities remain dependent on both
the target model and the safety scenario.

\paragraph{Response Perplexity}
Table~\ref{table:white PPL} further evaluates
the PPL of generated responses. JC achieves
average PPL values of 5.07, 4.57, and 5.07
on Qwen3-VL-8B-Instruct, InternVL3.5-8B,
and Gemma-3-12B-IT, respectively, compared
with 7.16, 6.79, and 6.86 for Plain Text.
JC also consistently achieves lower average
PPL than the two visual jailbreak baselines
across all three models. Moreover, JC obtains
the lowest PPL in 11 out of 13 scenarios
for each model, indicating that its generated
responses generally receive higher likelihood
under the evaluation language model.
However, higher PPL is observed in certain
scenarios, particularly Government Decision,
where JC exhibits increased PPL across all
three models. These findings indicate that
JC can achieve high attack effectiveness
while maintaining favorable
response likelihood.

\begin{table*}[t]
\centering
\small
\caption{White-box ASR (\%) of different jailbreak methods
across 13 safety scenarios on three MLLMs.
Results are reported with their corresponding
uncertainty estimates. Best results within each
model and scenario are highlighted in bold.
For brevity, column headers omit the
\mbox{-Instruct} and \mbox{-IT} suffixes of the model names.}
\label{table:white ASR}

\resizebox{\textwidth}{!}{
\begin{tabular}{lcccccccccccc}
\toprule
\multirow{2}{*}{\textbf{Scenario}} &
\multicolumn{4}{c}{\textbf{Qwen3-VL-8B}} &
\multicolumn{4}{c}{\textbf{InternVL3.5-8B}} &
\multicolumn{4}{c}{\textbf{Gemma-3-12B}} \\
\cmidrule(lr){2-5}
\cmidrule(lr){6-9}
\cmidrule(lr){10-13}
& PT & AE & QI & JC
& PT & AE & QI & JC
& PT & AE & QI & JC \\
\midrule
Economic Harm
& 4.37$\pm$2.00 & 17.07$\pm$4.42 & 14.59$\pm$3.59 & \textbf{100.00$\pm$0.00} 
& 4.23$\pm$1.76 & 36.10$\pm$4.64 & 41.49$\pm$4.53 & \textbf{100.00$\pm$0.00} 
& 7.05$\pm$2.41 & 31.45$\pm$3.72 & 42.14$\pm$5.07 & \textbf{100.00$\pm$0.00} \\
Financial Advice 
& 9.56$\pm$2.50 & 15.91$\pm$4.24 & 26.54$\pm$3.78 & \textbf{100.00$\pm$0.00} 
& 2.04$\pm$1.64 & 37.91$\pm$4.31 & 16.47$\pm$3.39 & \textbf{100.00$\pm$0.00} 
& 7.85$\pm$2.93 & 34.75$\pm$4.12 & 40.34$\pm$4.11 & \textbf{97.09$\pm$1.65} \\
Fraud 
& 7.59$\pm$2.64 & 15.97$\pm$3.81 & 29.51$\pm$5.11 & \textbf{57.54$\pm$10.90} 
& 8.77$\pm$2.87 & 30.33$\pm$3.90 & 22.59$\pm$3.41 & \textbf{62.76$\pm$10.60} 
& 3.14$\pm$2.02 & 9.63$\pm$3.01 & 16.80$\pm$3.54 & \textbf{96.79$\pm$0.51} \\
Government Decision 
& 6.39$\pm$1.99 & 20.56$\pm$4.72 & 8.77$\pm$2.35 & \textbf{41.19$\pm$10.80} 
& 6.61$\pm$2.79 & 38.30$\pm$4.55 & 35.71$\pm$5.00 & \textbf{94.98$\pm$1.96} 
& 7.55$\pm$3.05 & 28.46$\pm$4.86 & 32.08$\pm$4.23 & \textbf{100.00$\pm$0.00} \\
Hate Speech 
& 2.40$\pm$1.78 & 28.94$\pm$3.79 & 30.09$\pm$5.48 & \textbf{85.85$\pm$8.57} 
& 3.98$\pm$2.26 & 27.76$\pm$5.27 & 15.69$\pm$3.64 & \textbf{100.00$\pm$0.00} 
& 4.31$\pm$2.26 & 17.61$\pm$4.04 & 38.05$\pm$4.36 & \textbf{100.00$\pm$0.00} \\
Health Consultation 
& 2.40$\pm$1.38 & 25.41$\pm$3.82 & 13.48$\pm$3.79 & \textbf{100.00$\pm$0.00} 
& 1.57$\pm$1.15 & 28.86$\pm$4.21 & 9.93$\pm$2.45 & \textbf{100.00$\pm$0.00} 
& 6.69$\pm$2.64 & 33.52$\pm$4.20 & 28.10$\pm$5.35 & \textbf{100.00$\pm$0.00} \\
Illegal Activity 
& 1.52$\pm$1.30 & 20.07$\pm$3.28 & 9.47$\pm$2.58 & \textbf{82.23$\pm$7.64} 
& 3.80$\pm$1.61 & 25.25$\pm$4.38 & 18.01$\pm$3.50 & \textbf{67.51$\pm$10.29} 
& 6.70$\pm$2.08 & 15.63$\pm$3.94 & 27.13$\pm$4.26 & \textbf{100.00$\pm$0.00} \\
Legal Opinion 
& 8.80$\pm$2.62 & 32.25$\pm$4.35 & 43.06$\pm$3.97 & \textbf{82.55$\pm$8.57} 
& 3.93$\pm$1.73 & 9.97$\pm$3.24 & 13.13$\pm$3.93 & \textbf{90.50$\pm$3.40} 
& 5.82$\pm$2.02 & 35.26$\pm$4.28 & 16.19$\pm$4.26 & \textbf{96.99$\pm$1.65} \\
Malware Generation 
& 6.41$\pm$2.47 & 14.30$\pm$3.34 & 43.69$\pm$5.59 & \textbf{88.91$\pm$7.64} 
& 7.57$\pm$2.57 & 13.10$\pm$3.19 & \textbf{42.33$\pm$4.43} & 9.59$\pm$1.54 
& 1.81$\pm$1.55 & 23.70$\pm$3.96 & 10.54$\pm$3.23 & \textbf{100.00$\pm$0.00} \\
Physical Harm 
& 7.37$\pm$2.66 & 20.10$\pm$3.64 & 37.72$\pm$5.25 & \textbf{81.38$\pm$5.75} 
& 6.74$\pm$2.83 & 10.19$\pm$3.60 & 37.71$\pm$4.16 & \textbf{91.49$\pm$8.40} 
& 8.52$\pm$2.91 & 44.60$\pm$5.46 & 41.09$\pm$5.50 & \textbf{100.00$\pm$0.00} \\
Political Lobbying 
& 1.19$\pm$0.95 & 22.92$\pm$4.76 & 18.58$\pm$4.25 & \textbf{100.00$\pm$0.00} 
& 8.98$\pm$3.27 & 33.18$\pm$5.58 & 31.07$\pm$4.61 & \textbf{100.00$\pm$0.00} 
& 3.89$\pm$1.55 & 14.23$\pm$3.70 & 41.22$\pm$4.93 & \textbf{100.00$\pm$0.00} \\
Privacy Violence 
& 9.73$\pm$3.52 & 26.33$\pm$4.15 & 10.71$\pm$3.43 & \textbf{72.54$\pm$9.86} 
& 5.25$\pm$1.79 & 20.95$\pm$3.67 & 40.12$\pm$5.85 & \textbf{98.14$\pm$1.23} 
& 2.68$\pm$1.36 & 21.96$\pm$4.72 & 31.06$\pm$4.77 & \textbf{82.44$\pm$7.64} \\
Sex 
& 8.49$\pm$3.09 & 38.84$\pm$4.45 & 33.00$\pm$3.90 & \textbf{100.00$\pm$0.00} 
& 2.08$\pm$1.43 & 28.33$\pm$4.50 & 37.54$\pm$4.34 & \textbf{100.00$\pm$0.00} 
& 1.37$\pm$1.24 & 13.31$\pm$3.61 & 19.88$\pm$3.98 & \textbf{100.00$\pm$0.00} \\
\midrule
\textbf{Average}
& 5.86$\pm$0.56 & 22.97$\pm$0.88 & 24.55$\pm$0.93 & \textbf{84.01$\pm$1.61} 
& 5.04$\pm$0.48 & 26.17$\pm$1.09 & 27.83$\pm$0.90 & \textbf{85.77$\pm$1.05} 
& 5.18$\pm$0.48 & 24.93$\pm$0.90 & 29.59$\pm$1.03 & \textbf{97.95$\pm$0.49} \\
\bottomrule
\end{tabular}
}
\begin{flushleft}
\footnotesize
PT: Plain Text; AE: Adv Example;
QI: Query Image; JC: Jailbreak Connectivity.
\end{flushleft}
\end{table*}

\begin{table*}[t]
\centering
\small
\caption{PPL of generated responses under different
jailbreak methods across 13 safety scenarios.
Lower PPL indicates higher likelihood under the
evaluation language model. Best results within each
model and scenario are highlighted in bold.}
\label{table:white PPL}

\resizebox{\textwidth}{!}{
\begin{tabular}{lcccccccccccc}
\toprule
\multirow{2}{*}{\textbf{Scenario}} &
\multicolumn{4}{c}{\textbf{Qwen3-VL-8B}} &
\multicolumn{4}{c}{\textbf{InternVL3.5-8B}} &
\multicolumn{4}{c}{\textbf{Gemma-3-12B}} \\
\cmidrule(lr){2-5}
\cmidrule(lr){6-9}
\cmidrule(lr){10-13}
& PT & AE & QI & JC
& PT & AE & QI & JC
& PT & AE & QI & JC \\
\midrule
Economic Harm 
& 6.50$\pm$0.89 & 15.31$\pm$1.79 & 24.39$\pm$1.08 & \textbf{2.12$\pm$0.14} 
& 6.43$\pm$0.44 & 27.83$\pm$0.85 & 47.04$\pm$0.64 & \textbf{2.75$\pm$0.21} 
& 7.69$\pm$0.73 & 24.77$\pm$0.11 & 47.59$\pm$1.12 & \textbf{3.33$\pm$0.37} \\
Financial Advice 
& 8.80$\pm$0.32 & 14.55$\pm$1.71 & 34.46$\pm$0.35 & \textbf{3.13$\pm$0.34} 
& 5.46$\pm$1.71 & 29.02$\pm$0.50 & 25.98$\pm$0.61 & \textbf{2.90$\pm$0.40} 
& 8.05$\pm$1.41 & 26.94$\pm$0.39 & 46.07$\pm$0.27 & \textbf{3.95$\pm$0.48} \\
Fraud 
& 7.93$\pm$0.99 & 14.59$\pm$1.18 & 36.96$\pm$1.54 & \textbf{2.34$\pm$0.35} 
& 8.45$\pm$1.07 & 24.03$\pm$0.32 & 31.13$\pm$0.17 & \textbf{2.93$\pm$0.67} 
& 5.95$\pm$1.71 & 10.41$\pm$1.09 & 26.25$\pm$0.76 & \textbf{4.35$\pm$0.58} \\
Government Decision 
& \textbf{7.39$\pm$0.16} & 17.61$\pm$1.76 & 19.49$\pm$0.23 & 26.26$\pm$4.11 
& \textbf{7.49$\pm$1.55} & 29.27$\pm$0.71 & 42.18$\pm$1.20 & 20.67$\pm$5.73 
& \textbf{7.91$\pm$1.70} & 22.80$\pm$1.35 & 39.12$\pm$0.58 & 16.49$\pm$1.90 \\
Hate Speech 
& 5.62$\pm$1.74 & 23.12$\pm$0.26 & 37.44$\pm$1.88 & \textbf{2.32$\pm$0.52} 
& 6.32$\pm$1.71 & 22.34$\pm$1.80 & 25.32$\pm$1.00 & \textbf{2.99$\pm$0.40} 
& 6.47$\pm$1.50 & 15.66$\pm$1.27 & 44.15$\pm$0.54 & \textbf{4.20$\pm$0.86} \\
Health Consultation 
& \textbf{5.62$\pm$0.57} & 20.80$\pm$0.45 & 23.46$\pm$1.49 & 11.40$\pm$2.02 
& 5.25$\pm$0.67 & 23.07$\pm$0.68 & 20.46$\pm$0.19 & \textbf{2.04$\pm$0.17} 
& \textbf{7.53$\pm$1.26} & 26.13$\pm$0.50 & 35.77$\pm$1.85 & 7.62$\pm$3.08 \\
Illegal Activity 
& 5.23$\pm$1.29 & 17.28$\pm$0.18 & 20.08$\pm$0.46 & \textbf{2.31$\pm$0.27} 
& 6.24$\pm$0.30 & 20.69$\pm$1.03 & 27.27$\pm$0.60 & \textbf{2.98$\pm$0.16} 
& 7.53$\pm$0.25 & 14.36$\pm$1.38 & 34.95$\pm$0.81 & \textbf{3.01$\pm$0.38} \\
Legal Opinion 
& 8.46$\pm$0.66 & 25.30$\pm$0.69 & 48.36$\pm$0.11 & \textbf{2.77$\pm$0.85} 
& \textbf{6.30$\pm$0.51} & 10.64$\pm$1.37 & 23.16$\pm$1.73 & 8.43$\pm$2.40 
& 7.14$\pm$0.39 & 27.27$\pm$0.53 & 25.74$\pm$1.71 & \textbf{4.67$\pm$0.95} \\
Malware Generation 
& 7.40$\pm$1.04 & 13.49$\pm$0.80 & 48.90$\pm$1.57 & \textbf{2.26$\pm$0.41} 
& 7.92$\pm$0.87 & 12.70$\pm$0.75 & 47.75$\pm$0.53 & \textbf{2.95$\pm$0.22} 
& 5.36$\pm$1.72 & 19.67$\pm$0.69 & 20.98$\pm$1.24 & \textbf{3.61$\pm$0.42} \\
Physical Harm 
& 7.83$\pm$1.08 & 17.30$\pm$0.59 & 43.87$\pm$1.37 & \textbf{2.99$\pm$0.34} 
& 7.55$\pm$1.57 & 10.79$\pm$1.85 & 43.86$\pm$0.36 & \textbf{2.85$\pm$0.39} 
& 8.34$\pm$1.19 & 33.42$\pm$1.44 & 46.71$\pm$1.53 & \textbf{3.37$\pm$0.46} \\
Political Lobbying 
& 5.08$\pm$0.43 & 19.16$\pm$1.59 & 27.75$\pm$1.41 & \textbf{2.43$\pm$0.14} 
& 8.55$\pm$1.65 & 25.91$\pm$1.83 & 38.27$\pm$0.98 & \textbf{2.47$\pm$0.26} 
& 6.28$\pm$0.12 & 13.44$\pm$1.27 & 46.81$\pm$1.00 & \textbf{3.81$\pm$0.66} \\
Privacy Violence 
& 8.88$\pm$1.85 & 21.40$\pm$0.74 & 21.13$\pm$1.49 & \textbf{1.91$\pm$0.15} 
& 6.89$\pm$0.11 & 17.86$\pm$0.55 & 45.89$\pm$1.87 & \textbf{2.77$\pm$0.35} 
& 5.75$\pm$0.28 & 18.53$\pm$1.63 & 38.26$\pm$1.14 & \textbf{3.60$\pm$0.31} \\
Sex 
& 8.33$\pm$1.50 & 29.63$\pm$0.61 & 39.90$\pm$0.23 & \textbf{3.67$\pm$0.14} 
& 5.48$\pm$1.02 & 22.71$\pm$1.00 & 43.72$\pm$0.54 & \textbf{2.73$\pm$0.27} 
& 5.16$\pm$1.29 & 12.84$\pm$1.28 & 28.85$\pm$0.99 & \textbf{3.86$\pm$1.30} \\
\midrule
\textbf{Average} 
& 7.16$\pm$0.28 & 19.20$\pm$0.25 & 32.78$\pm$0.25 & \textbf{5.07$\pm$0.34} 
& 6.79$\pm$0.30 & 21.30$\pm$0.26 & 35.54$\pm$0.20 & \textbf{4.57$\pm$0.38} 
& 6.86$\pm$0.27 & 20.48$\pm$0.26 & 37.02$\pm$0.25 & \textbf{5.07$\pm$0.28} \\
\bottomrule
\end{tabular}
}
\end{table*}

\subsubsection{Transferable Jailbreak Performance}

We evaluate the cross-model transferability of JC
using Qwen3-VL-8B-Instruct as the source MLLM,
with InternVL3.5-8B and Gemma-3-12B-IT serving
as transfer targets. We compare JC against
FORCE~\cite{lin2026force}, a recent visual
jailbreak method designed to improve attack
transferability by mitigating model-specific
feature reliance through layer-aware regularization
and spectral feature rescaling.

We evaluate both methods across 13 safety
scenarios under a consistent evaluation protocol.
Table~\ref{table:transferable} reports their
ASR on the source model and the two transfer
targets.

\begin{table}[htbp]
\centering
\small
\caption{Transferable jailbreak ASR (\%) of FORCE
and JC across 13 safety scenarios. Adversarial
images are generated using Qwen3-VL-8B-Instruct
as the source model and evaluated on
InternVL3.5-8B and Gemma-3-12B-IT as transfer
targets. Best results are highlighted in bold.}
\label{table:transferable}

\resizebox{\linewidth}{!}{
\begin{tabular}{lcccccc}
\toprule
\multirow{2}{*}{\textbf{Scenario}} &
\multicolumn{2}{c}{\textbf{Qwen3-VL-8B (Source)}} &
\multicolumn{2}{c}{\textbf{InternVL3.5-8B}} &
\multicolumn{2}{c}{\textbf{Gemma-3-12B}} \\
\cmidrule(lr){2-3}
\cmidrule(lr){4-5}
\cmidrule(lr){6-7}
& FORCE & JC & FORCE & JC & FORCE & JC \\
\midrule
Economic Harm
& 84.24$\pm$3.21 & \textbf{87.12$\pm$3.27}
& \textbf{82.09$\pm$3.62} & 80.60$\pm$4.29
& \textbf{81.61$\pm$3.75} & 75.61$\pm$3.95 \\

Financial Advice
& \textbf{92.31$\pm$2.68} & 86.82$\pm$2.87
& \textbf{81.42$\pm$3.29} & 74.97$\pm$4.78
& 69.65$\pm$4.09 & \textbf{83.01$\pm$3.43} \\

Fraud
& \textbf{51.35$\pm$5.18} & 49.96$\pm$4.99
& 48.31$\pm$5.72 & \textbf{52.86$\pm$5.11}
& 71.78$\pm$3.82 & \textbf{77.97$\pm$3.38} \\

Government Decision
& 35.99$\pm$3.93 & \textbf{36.26$\pm$3.91}
& \textbf{77.72$\pm$4.37} & 70.33$\pm$5.06
& 79.33$\pm$3.79 & \textbf{84.30$\pm$3.80} \\

Hate Speech
& 69.70$\pm$4.79 & \textbf{77.48$\pm$4.86}
& 69.33$\pm$4.30 & \textbf{86.82$\pm$3.37}
& 71.43$\pm$5.32 & \textbf{85.55$\pm$3.52} \\

Health Consultation
& 81.18$\pm$3.39 & \textbf{89.32$\pm$2.79}
& 70.94$\pm$3.75 & \textbf{83.81$\pm$3.72}
& 69.15$\pm$4.29 & \textbf{77.45$\pm$3.43} \\

Illegal Activity
& 65.63$\pm$3.92 & \textbf{72.29$\pm$4.77}
& 46.36$\pm$4.61 & \textbf{51.16$\pm$4.85}
& 72.35$\pm$4.51 & \textbf{74.54$\pm$3.97} \\

Legal Opinion
& \textbf{75.22$\pm$5.09} & \textbf{75.22$\pm$3.99}
& 65.96$\pm$4.41 & \textbf{66.13$\pm$3.83}
& 68.30$\pm$5.03 & \textbf{73.90$\pm$5.11} \\

Malware Generation
& \textbf{77.72$\pm$4.94} & 76.81$\pm$4.25
& 27.08$\pm$2.80 & \textbf{28.10$\pm$2.30}
& \textbf{81.95$\pm$3.64} & 78.98$\pm$3.65 \\

Physical Harm
& \textbf{72.36$\pm$5.02} & 71.55$\pm$4.60
& 65.94$\pm$5.00 & \textbf{75.84$\pm$3.48}
& 80.12$\pm$4.74 & \textbf{84.45$\pm$3.11} \\

Political Lobbying
& 79.29$\pm$3.74 & \textbf{88.66$\pm$2.77}
& 80.43$\pm$4.58 & \textbf{83.21$\pm$3.94}
& 77.50$\pm$4.95 & \textbf{85.05$\pm$3.55} \\

Privacy Violence
& \textbf{67.16$\pm$3.94} & 64.97$\pm$5.67
& 71.99$\pm$4.44 & \textbf{82.24$\pm$3.54}
& \textbf{66.84$\pm$4.24} & 60.26$\pm$5.84 \\

Sex
& 90.65$\pm$3.13 & \textbf{92.85$\pm$2.86}
& 72.21$\pm$3.80 & \textbf{74.04$\pm$4.40}
& 80.06$\pm$3.99 & \textbf{80.15$\pm$3.58} \\
\midrule
\textbf{Average} 
& 72.52$\pm$1.02 & \textbf{74.56$\pm$0.88} 
& 66.14$\pm$0.90 & \textbf{70.01$\pm$0.91} 
& 74.62$\pm$0.97 & \textbf{78.56$\pm$0.90} \\
\bottomrule
\end{tabular}
}
\end{table}

The transferability of JC nevertheless varies
across safety scenarios. On InternVL3.5-8B,
JC achieves particularly notable improvements
in Hate Speech, Health Consultation, and
Privacy Violence, with ASR gains of 17.49,
12.87, and 10.25 percentage points, respectively.
On Gemma-3-12B-IT, JC improves ASR by 14.12
percentage points in Hate Speech and 13.36
percentage points in Financial Advice.
However, FORCE performs better in several
categories, including Economic Harm on both
transfer targets and Privacy Violence on
Gemma-3-12B-IT. These findings indicate that
JC provides stronger transferability on average,
while its effectiveness remains dependent
on the target model and safety scenario. A detailed analysis of surrogate fidelity and hyperparameter sensitivity is provided in Sections~\ref{S-appendix:surrogate_analysis} and~\ref{S-appendix:hyperparameter} of the supplementary material, respectively.

\subsubsection{Black-Box Diverse Attacks}

We further evaluate the effectiveness of JC
against two proprietary MLLMs, GPT-5.6 and
Gemini 3.1 Pro, under a black-box setting.
Specifically, we use Qwen3-VL-8B-Instruct
as the base model to generate adversarial
jailbreak images, which are subsequently
evaluated on the two target MLLMs without
accessing their model parameters or gradients.
We compare JC with FigStep~\cite{gong2025figstep},
a representative visual jailbreak baseline.
Both methods are evaluated across the same
13 safety scenarios. Table~\ref{table:black}
reports the ASR and PPL, expressed as
mean $\pm$ standard deviation over five
independent runs.

\paragraph{Attack Effectiveness}
As shown in Table~\ref{table:black},
JC substantially outperforms FigStep
in terms of ASR across both proprietary
MLLMs. On GPT-5.6, JC achieves an average
ASR of 41.94\%, compared with 17.25\%
for FigStep, yielding an improvement of
24.69 percentage points. Similarly,
on Gemini 3.1 Pro, JC increases the
average ASR from 19.71\% to 47.05\%,
representing a gain of 27.34 percentage
points. Notably, JC consistently
outperforms FigStep across all 13
safety scenarios on both models.

The improvements are particularly
pronounced in certain categories.
On GPT-5.6, JC achieves ASRs of 60.68\%
in Privacy Violence, 59.83\% in
Financial Advice, and 56.11\% in
Legal Opinion. On Gemini 3.1 Pro,
JC achieves 67.46\% in Economic Harm,
65.58\% in Financial Advice, and
62.81\% in Political Lobbying.
These results demonstrate that the
adversarial images generated by JC
can transfer to proprietary MLLMs
and achieve higher attack success
rates than FigStep, although the
effectiveness varies across models
and safety scenarios.

\paragraph{Response Perplexity}
We further examine the PPL of generated
responses under the evaluation language
model. JC obtains average PPL values of
11.41 and 13.11 on GPT-5.6 and Gemini 3.1
Pro, respectively, lower than the 14.52 and
16.41 obtained by FigStep. Moreover, JC
yields lower PPL than FigStep in all 13
safety scenarios on both proprietary MLLMs.
These results indicate that, in the black-box
setting, JC attains both higher attack success
rates and more fluent responses than FigStep.

\begin{table*}[t]
\centering
\small
\caption{
Black-box jailbreak performance of FigStep
and JC against GPT-5.6 and Gemini 3.1 Pro
across 13 safety scenarios.
Qwen3-VL-8B-Instruct serves as the base
model for JC. Results are reported as
mean $\pm$ standard deviation over
five independent runs. Higher ASR
and lower PPL are preferable.
}
\label{table:black}

\resizebox{\textwidth}{!}{
\begin{tabular}{lcccccccc}
\toprule
\multirow{3}{*}{\textbf{Scenario}} &
\multicolumn{4}{c}{\textbf{GPT-5.6}} &
\multicolumn{4}{c}{\textbf{Gemini 3.1 Pro}} \\
\cmidrule(lr){2-5}
\cmidrule(lr){6-9}
& \multicolumn{2}{c}{\textbf{ASR (\%) $\uparrow$}}
& \multicolumn{2}{c}{\textbf{PPL $\downarrow$}}
& \multicolumn{2}{c}{\textbf{ASR (\%) $\uparrow$}}
& \multicolumn{2}{c}{\textbf{PPL $\downarrow$}} \\
\cmidrule(lr){2-3}
\cmidrule(lr){4-5}
\cmidrule(lr){6-7}
\cmidrule(lr){8-9}
& FigStep & JC & FigStep & JC
& FigStep & JC & FigStep & JC \\
\midrule
Economic Harm 
& 12.74$\pm$2.93 & \textbf{34.48$\pm$4.18} & 11.06$\pm$1.89 & \textbf{9.55$\pm$0.89} 
& 33.25$\pm$4.44 & \textbf{67.46$\pm$4.42} & 16.90$\pm$1.02 & \textbf{13.51$\pm$1.38} \\
Financial Advice 
& 27.72$\pm$4.50 & \textbf{59.83$\pm$4.93} & 12.03$\pm$0.24 & \textbf{9.18$\pm$0.32} 
& 31.95$\pm$3.95 & \textbf{65.58$\pm$4.02} & 12.33$\pm$0.82 & \textbf{8.69$\pm$1.47} \\
Fraud 
& 22.03$\pm$4.30 & \textbf{50.21$\pm$5.18} & 10.97$\pm$0.90 & \textbf{9.20$\pm$0.99} 
& 23.34$\pm$4.84 & \textbf{53.11$\pm$5.72} & 19.35$\pm$0.24 & \textbf{16.63$\pm$1.11} \\
Government Decision 
& 18.57$\pm$3.18 & \textbf{44.34$\pm$4.07} & 14.15$\pm$0.29 & \textbf{10.65$\pm$0.16} 
& 32.73$\pm$4.92 & \textbf{56.72$\pm$4.94} & 9.41$\pm$0.53 & \textbf{7.89$\pm$1.49} \\
Hate Speech 
& 9.67$\pm$2.67 & \textbf{27.06$\pm$4.51} & 17.84$\pm$1.83 & \textbf{13.30$\pm$1.74} 
& 8.57$\pm$2.61 & \textbf{31.72$\pm$4.34} & 21.68$\pm$0.32 & \textbf{18.84$\pm$0.99} \\
Health Consultation 
& 7.06$\pm$2.22 & \textbf{24.86$\pm$3.75} & 16.59$\pm$0.11 & \textbf{12.18$\pm$0.57} 
& 11.68$\pm$2.65 & \textbf{36.23$\pm$3.97} & 18.67$\pm$0.43 & \textbf{16.27$\pm$1.04} \\
Illegal Activity 
& 4.51$\pm$1.71 & \textbf{20.56$\pm$3.34} & 13.19$\pm$1.49 & \textbf{10.49$\pm$1.29} 
& 7.31$\pm$2.41 & \textbf{29.90$\pm$4.23} & 13.17$\pm$0.53 & \textbf{9.38$\pm$0.87} \\
Legal Opinion 
& 25.52$\pm$5.14 & \textbf{56.11$\pm$5.85} & 16.19$\pm$1.57 & \textbf{14.34$\pm$0.66} 
& 15.43$\pm$3.36 & \textbf{41.66$\pm$4.59} & 9.21$\pm$0.85 & \textbf{7.07$\pm$0.15} \\
Malware Generation 
& 18.63$\pm$4.62 & \textbf{44.45$\pm$5.89} & 10.23$\pm$1.66 & \textbf{8.67$\pm$1.04} 
& 17.27$\pm$4.13 & \textbf{44.32$\pm$5.42} & 21.70$\pm$0.19 & \textbf{16.79$\pm$0.29} \\
Physical Harm 
& 21.41$\pm$4.61 & \textbf{49.16$\pm$5.62} & 14.92$\pm$0.25 & \textbf{10.51$\pm$1.08} 
& 13.87$\pm$3.65 & \textbf{39.40$\pm$5.16} & 17.81$\pm$1.72 & \textbf{15.48$\pm$0.16} \\
Political Lobbying 
& 3.54$\pm$1.70 & \textbf{18.91$\pm$3.61} & 13.15$\pm$0.82 & \textbf{11.40$\pm$0.43} 
& 30.03$\pm$5.29 & \textbf{62.81$\pm$5.58} & 22.03$\pm$1.80 & \textbf{15.75$\pm$1.25} \\
Privacy Violence 
& 28.22$\pm$3.78 & \textbf{60.68$\pm$4.10} & 17.32$\pm$0.57 & \textbf{12.47$\pm$1.85} 
& 16.35$\pm$3.66 & \textbf{42.98$\pm$4.90} & 21.43$\pm$0.98 & \textbf{16.26$\pm$0.67} \\
Sex 
& 24.64$\pm$4.63 & \textbf{54.63$\pm$5.35} & 21.16$\pm$1.54 & \textbf{16.42$\pm$1.50} 
& 14.49$\pm$2.96 & \textbf{39.80$\pm$4.15} & 9.65$\pm$0.98 & \textbf{7.89$\pm$1.02} \\
\midrule
\textbf{Average} 
& 17.25$\pm$0.87 & \textbf{41.94$\pm$0.98} & 14.52$\pm$0.26 & \textbf{11.41$\pm$0.27} 
& 19.71$\pm$0.79 & \textbf{47.05$\pm$1.23} & 16.41$\pm$0.23 & \textbf{13.11$\pm$0.24} \\
\bottomrule
\end{tabular}
}
\end{table*}

\subsubsection{Universal Attacks}
To evaluate the universal attack capability of JC, we tested on 40 unseen harmful queries and compared with the Adv Example baseline~\cite{qi2024visual}. Averaged over five independent runs, JC achieves 34/40 successful jailbreaks on Qwen3-VL-8B-Instruct, significantly outperforming the baseline (17/40). Moreover, when transferred to InternVL3.5-8B, the same universal attack image remains effective on 22 queries on average, confirming cross-model transferability. Additional analysis on the impact of harmful corpus construction is provided in Section~\ref{S-appendix:transfer_additional_experiments} of the supplementary material.

\subsubsection{Ablation Study}
\label{sec:ablation_study}

To investigate the contribution of each component
in JC, we conduct ablation experiments on three
MLLMs: Qwen3-VL-8B-Instruct, InternVL3.5-8B,
and Gemma-3-12B-IT. Specifically, we consider
three variants by individually removing
(i) the continuous path construction
(\textbf{w/o Path}),
(ii) the Jailbreak Success Predictor
(\textbf{w/o JSP}), and
(iii) the Safety Classifier
(\textbf{w/o SC}). We perform the ablation study under the
transferable jailbreak setting, using
Qwen3-VL-8B-Instruct as the source model
and InternVL3.5-8B and Gemma-3-12B-IT
as transfer targets.
All other experimental settings remain
consistent with the full JC method.
Table~\ref{table:jc-ablation} reports
the average ASR and PPL across 13 safety
scenarios.

\begin{table}[t]
\centering
\small
\caption{
Ablation study of JC under the transferable
jailbreak setting, with Qwen3-VL-8B-Instruct
as the source model and InternVL3.5-8B and
Gemma-3-12B-IT as transfer targets. We report
ASR (\%) and PPL as mean $\pm$ standard deviation
over five independent runs. Higher ASR and lower
PPL are preferable. Best results are highlighted
in bold.
}
\label{table:jc-ablation}

\resizebox{\linewidth}{!}{
\begin{tabular}{lcccccc}
\toprule
\multirow{2}{*}{\textbf{Configuration}} &
\multicolumn{2}{c}{\textbf{Qwen3-VL-8B-Instruct}} &
\multicolumn{2}{c}{\textbf{InternVL3.5-8B}} &
\multicolumn{2}{c}{\textbf{Gemma-3-12B-IT}} \\
\cmidrule(lr){2-3}
\cmidrule(lr){4-5}
\cmidrule(lr){6-7}
& \textbf{ASR (\%) $\uparrow$}
& \textbf{PPL $\downarrow$}
& \textbf{ASR (\%) $\uparrow$}
& \textbf{PPL $\downarrow$}
& \textbf{ASR (\%) $\uparrow$}
& \textbf{PPL $\downarrow$} \\
\midrule
JC (Full) 
& 74.56$\pm$0.88 & \textbf{5.84$\pm$0.36} 
& \textbf{70.01$\pm$0.91} & \textbf{3.46$\pm$0.49} 
& \textbf{78.56$\pm$0.90} & \textbf{5.77$\pm$0.32} \\
w/o Path 
& 28.65$\pm$4.49 & 7.94$\pm$1.08 
& 27.68$\pm$3.19 & 4.82$\pm$0.90 
& 35.04$\pm$2.96 & 7.84$\pm$1.16 \\
w/o JSP 
& \textbf{79.34$\pm$3.30} & 7.44$\pm$1.11 
& 54.85$\pm$4.68 & 3.83$\pm$0.82 
& 54.04$\pm$4.58 & 7.25$\pm$1.29 \\
w/o SC 
& 75.85$\pm$3.40 & 8.31$\pm$1.64 
& 39.32$\pm$2.58 & 4.08$\pm$1.22 
& 55.62$\pm$5.23 & 8.51$\pm$1.31 \\
\bottomrule
\end{tabular}
}
\end{table}

\paragraph{Contribution of Individual Components}
As shown in Table~\ref{table:jc-ablation},
removing the continuous path construction
causes the largest and most consistent
degradation, reducing the transfer ASR by
45.91, 42.33, and 43.52 percentage points on
Qwen3-VL-8B-Instruct, InternVL3.5-8B, and
Gemma-3-12B-IT, respectively. This confirms
that continuous path-based optimization is the
primary driver of effective jailbreaks.

The two surrogate classifiers act mainly on
the transfer targets rather than on the source
model. Removing the JSP reduces the ASR by
15.16 and 24.52 percentage points on the two
transfer targets (InternVL3.5-8B and
Gemma-3-12B-IT), and removing the SC reduces it
by 30.69 and 22.94 percentage points,
respectively. On the source model
(Qwen3-VL-8B-Instruct), where no surrogate
guidance is applied, removing either classifier
leaves the source ASR essentially unchanged (a
slight increase of 4.78 for w/o JSP and 1.29
for w/o SC), since the freed optimization
capacity is spent on the source objective at
the cost of transferability. These results
confirm that the SC and JSP are complementary
and are the key contributors to cross-model
transfer.

\paragraph{Impact on Response Perplexity}
The full JC method attains the lowest PPL among
all configurations on every model, with values
of 5.84, 3.46, and 5.77 on Qwen3-VL-8B-Instruct,
InternVL3.5-8B, and Gemma-3-12B-IT, respectively.
Removing any component increases the PPL; for
example, the PPL rises to 8.31 and 8.51 without
the SC on Qwen3-VL-8B-Instruct and Gemma-3-12B-IT,
respectively. This indicates that the complete
design not only maximizes attack success but also
yields the most fluent responses.

Overall, the ablation results show that the
continuous path drives attack effectiveness,
the two surrogate classifiers drive cross-model
transferability, and all three components jointly
yield the lowest response perplexity.

\section{Conclusion}
In this paper, we proposed JC, a unified framework for visual jailbreak attacks on MLLMs, grounded in the insight that adversarial effectiveness forms a connected low-loss region in image space. By constructing a continuous path between two distinct jailbreak endpoints, JC unifies three capabilities: diverse jailbreak generation via path sampling, improved transferability through surrogate-guided optimization, and universal attacks by extending the objective to a harmful output distribution. Experiments show that JC significantly outperforms prior methods in attack success rate, while attaining lower response perplexity than the baselines in both white-box and black-box settings. We note, however, that perplexity only reflects fluency under a fixed evaluation model and does not directly measure harmfulness. Our findings suggest that safety alignment in MLLMs may fail at the region level, rather than at isolated inputs, indicating that defenses must reason about structured adversarial regions rather than individual examples.

%\section*{Acknowledgments}
%This should be a simple paragraph before the References to thank those individuals and institutions who have supported your work on this article.

%{\appendix[Proof of the Zonklar Equations]
%Use $\backslash${\tt{appendix}} if you have a single appendix:
%Do not use $\backslash${\tt{section}} anymore after $\backslash${\tt{appendix}}, only $\backslash${\tt{section*}}.
%If you have multiple appendixes use $\backslash${\tt{appendices}} then use $\backslash${\tt{section}} to start each appendix.
%You must declare a $\backslash${\tt{section}} before using any $\backslash${\tt{subsection}} or using $\backslash${\tt{label}} ($\backslash${\tt{appendices}} by itself
% starts a section numbered zero.)}


\bibliographystyle{IEEEtran}
\bibliography{ref}

%\newpage
% your own custom macro containing the $\backslash${\tt{includegraphics}} command to make things
%\bf{If you include a photo:}\vspace{-33pt}
%\begin{IEEEbiography}[{\includegraphics[width=1in,height=1.25in,clip,keepaspectratio]{fig1}}]{Michael Shell}
%Use $\backslash${\tt{begin\{IEEEbiography\}}} and then for the 1st argument use $\backslash${\tt{includegraphics}} to declare and link the author photo.
%Use the author name as the 3rd argument followed by the biography text.
%\end{IEEEbiography}

\vfill

\end{document}